\documentclass{ecai} 

\usepackage{graphicx}
\usepackage{latexsym}
\usepackage{booktabs}
\usepackage{multirow}
\usepackage{color}
\usepackage{bbding}
\usepackage{subcaption} 
\usepackage{caption}   
\usepackage{float}     
\usepackage{amsmath}   
\usepackage{amssymb}   
\usepackage{tabularx}   
\usepackage{multirow}
\usepackage[table,xcdraw]{xcolor}
\usepackage{colortbl}
\usepackage{makecell}  
\usepackage{colortbl} 
\usepackage{array}
\usepackage{booktabs}
\usepackage{threeparttable}

\usepackage{lineno,hyperref}

\newcommand{\BibTeX}{B\kern-.05em{\sc i\kern-.025em b}\kern-.08em\TeX}

\begin{document}


\begin{frontmatter}


\paperid{0370} 


\title{FTCFormer: Fuzzy Token Clustering Transformer for Image Classification}


\author[A]{\fnms{Muyi}~\snm{Bao}}
\author[A]{\fnms{Changyu}~\snm{Zeng}}
\author[A]{\fnms{Yifan}~\snm{Wang}} %
\author[B]{\fnms{Zhengni}~\snm{Yang}}
\author[A]{\fnms{Zimu}~\snm{Wang}}
\author[C]{\fnms{Guangliang}~\snm{Cheng}} 
\author[A]{\fnms{Jun}~\snm{Qi}}
\author[A]{\fnms{Wei}~\snm{Wang}\thanks{Corresponding Author. Email: Wei.Wang03@xjtlu.edu.cn}}

\address[A]{School of Advanced Technology, Xi'an Jiaotong-Liverpool University, Suzhou, China}
\address[B]{Department of Mathematical Sciences, University of Liverpool, Liverpool, United Kingdom}
\address[C]{Department of Computer Science, University of Liverpool, Liverpool, United Kingdom}

\begin{abstract}
Transformer-based deep neural networks have achieved remarkable success across various computer vision tasks, largely attributed to their long-range self-attention mechanism and scalability. However, most transformer architectures embed images into uniform, grid-based vision tokens, neglecting the underlying semantic meanings of image regions, resulting in suboptimal feature representations. To address this issue, we propose \textbf{\underline{F}}uzzy \textbf{\underline{T}}oken \textbf{\underline{C}}lustering Trans\textbf{\underline{former}} (\textbf{FTCFormer}), which incorporates a novel clustering-based downsampling module to dynamically generate vision tokens based on the semantic meanings instead of spatial positions. It allocates fewer tokens to less informative regions and more to represent semantically important regions, regardless of their spatial adjacency or shape irregularity. To further enhance feature extraction and representation, we propose a Density Peak Clustering-Fuzzy K-Nearest Neighbor (DPC-FKNN) mechanism for clustering center determination, a Spatial Connectivity Score (SCS) for token assignment, and a channel-wise merging (Cmerge) strategy for token merging. Extensive experiments on 32 datasets across diverse domains validate the effectiveness of FTCFormer on image classification, showing consistent improvements over the TCFormer baseline, achieving gains of improving 1.43\% on five fine-grained datasets, 1.09\% on six natural image datasets, 0.97\% on three medical datasets and 0.55\% on four remote sensing datasets. The code is available at: \href{https://github.com/BaoBao0926/FTCFormer/tree/main}{https://github.com/BaoBao0926/FTCFormer/tree/main}.
\end{abstract}

\end{frontmatter}

\section{Introduction}

Vision transformers (ViTs) have achieved significant progress in various computer vision (CV) tasks, such as image classification \cite{ViT, PVT, swin}, object detection \cite{DETR} and semantic segmentation \cite{segformer}. Leveraging the parallel self-attention and flexible scalability, ViTs are particularly effective at capturing long-range dependencies between image patches (tokens). To balance accuracy and computational complexity, many recent ViTs \cite{CVT, PVT, swin} adopt hierarchical architectures, which are typically constructed using conventional grid-based downsampling methods, such as MaxPooling, AvgPooling and strided CNNs. However, these fixed-shape tokenization strategies treat all image regions with equal importance and often disregard semantically meaningful regions \cite{tcformer, TCFormerv2}.




To address these issues, methods for dynamic token generation are proposed, which aim to generate vision tokens using semantic meaning instead of spatial positions for better feature representation. TCFormer \cite{tcformer, TCFormerv2} is the first work to explore this idea. It introduces a clustering-based downsampling module to generate non-fixed vision tokens (i.e. clustering and merging tokens) with the classic Density Peak Clustering-K Nearest Neighbor (DPC-KNN) algorithm. The novel design dynamically allocates fewer vision tokens to non-important image regions (e.g. sky in the background), while allocating more tokens to semantically important regions (e.g. human faces), enhancing feature extraction and representation.

Despite the superior performance, TCFormer also presents some notable limitations. First, in clustering center determination, the use of DPC-KNN is inherently sensitive to the $K$ value \cite{tcformer, dpcknn, IDPC-NNMS}. Moreover, it struggles with uneven data (e.g. learned semantic feature maps) \cite{DPC-FWSN, IDPC-NNMS, ToMe} and is vulnerable to outliers in the KNN set \cite{dpcknn, dpc}. In token assignment, only using Euclidean distance may suffer from the curse of dimensionality and overparameterized feature spaces that contain insignificant noise \cite{ToMe}. Finally, in token merging, it \cite{tcformer, TCFormerv2} regresses importance scores at the token level, disregarding semantic differences across individual channels within tokens. This can potentially lead to the loss of discriminative information and reduce the effectiveness of feature representation.

To tackle the issues, we propose these \textbf{\underline{F}}uzzy \textbf{\underline{T}}oken \textbf{\underline{C}}lustering Trans\textbf{\underline{former}} (\textbf{FTCFormer}), which employs the DPC-Fuzzy K Nearest Neighbor (DPC-FKNN) for clustering center determination, a new metric called Spatial Connectivity Score (SCS) for token assignment, and a channel-wise token merging strategy for merging tokens. Specifically, 
\textbf{(1)} Instead of only utilizing the KNN set, DPC-FKNN incorporates both the KNN set and distance-weighted FKNN set (tokens outside of the KNN set) to determine clustering centers, resulting in more effective handling of uncertainty and increased robustness to uneven data, $K$ values and noise. 
\textbf{(2)} To mitigate the limitation of relying solely on the Euclidean distance in token assignment, the proposed SCS metric jointly considers the number of Shared Nearest Neighbors (SNN) between tokens and their closeness to the KNN set. During token assignment, non-cluster-center tokens are first assigned using the SCS metric, while Euclidean distance is only used as the secondary criterion. 
\textbf{(3)} To preserve semantic diversity within feature channels, we propose Channel Merging (Cmerge), a strategy that regresses importance scores at the channel level rather than the token level. This approach retains finer-grained semantic information during the merging process, improving the overall quality of token representations. 
The components of DPC-FKNN, SCS, and Cmerge are integrated into the \textbf{\underline{F}}uzzy \textbf{\underline{T}}oken \textbf{\underline{C}}lustering and \textbf{\underline{M}}erging (\textbf{\underline{FTCM}}) module within the FTCFormer framework.

The main contributions can be summarized as follows:
\begin{itemize}
    \item We introduce the FTCFormer architecture, which incorporates a novel clustering-based token downsampling model to consolidate feature representations by allocating more vision tokens to semantically important areas while paying less attention to insignificant regions.
    
    \item We propose the Fuzzy Token Clustering and Merging (FTCM) module, which integrates three key components: DPC-FKNN for clustering center selection, SCS metric for token assignment, and channel merge (Cmerge) strategy for channel-level token merging, to support dynamic token generation.
    
    \item We conduct extensive experiments across 32 datasets, and demonstrate that FTCFormer consistently outperforms state-of-the-art models in image classification. Specifically, it achieves average accuracy improvements of 1.43\% on five fine-grained datasets, 1.09\% on six natural image datasets, 0.97\% on three medical image datasets, and 0.55\% on four remote sensing datasets, highlighting its robustness and generalization across diverse visual domains.
\end{itemize}

\section{Related Work}
\subsection{Dynamic Token Generation}

Dynamic token generation is first explored by the field of token pruning, a dynamic token reduction approach that identifies and processes redundant tokens to improve computational efficiency. Token pruning methods can be roughly categorized into two groups: (1) hard token pruning, which directly discards unimportant tokens, and (2) soft token pruning, which merges unimportant tokens into a compressed token.

Methods in the hard token pruning category directly discard unimportant tokens. They determine the importance of tokens by incorporating a prediction network \cite{DynamicVIT}, learning adaptive halting \cite{A-ViT} and employing interpretability-aware token selection \cite{IA-RED}.
As for soft token pruning methods, SPViT \cite{SPViT} and PnP-DETR \cite{PnP-DETR} integrate less important tokens into one token using a lightweight decision network. Evo-ViT \cite{EVO-ViT}, ATS \cite{ATS} and EViT \cite{EViT} distinguish informative tokens by using attention between class tokens and vision tokens. Later works utilize the bipartite soft matching with attention scores \cite{ToMe, ToFu, BAT}, diversity \cite{BAT} and multi-criterion \cite{MCTF}.

Token pruning methods, similar to downsampling layers, can significantly accelerate inference in non-hierarchical architectures, but at the expense of overall performance. In contrast, TCFormer \cite{tcformer} also aims at improving the capabilities of feature representation. It is the first work to explore this idea by employing DPC-KNN into a deep learning framework. TCFormerV2 \cite{TCFormerv2} clusters the vision tokens in the local window, significantly improving the efficiency. Based on TCFormer \cite{tcformer}, we refine each stage, i.e. clustering and merging processes, to improve performance in image classification tasks across various domains.

\subsection{Density Peak Clustering}

Clustering analysis is an unsupervised machine learning method that identifies inherent groupings in datasets without prior knowledge \citep{SFKNN-DPC, dpc}. The Density Peaks Clustering (DPC) algorithm \citep{dpc} is particularly effective for automated cluster detection without expensive iterative computation, based on two key assumptions that cluster centers should exhibit higher local density and own larger separation distance from other centers. Despite these advantages, DPC has limitations, such as sensitivity to hyperparameters and poor performance on unevenly distributed data \citep{dpcknn, AKE-DPC, DPC-BDFN, DPC-FSC, DPC-FWSN}, motivating several subsequent works for improvement.  

Most enhancements focus on redefining local density metrics and assignment strategies. 
For instance, DPC-KNN \citep{dpcknn} integrates KNN for better density estimation, while \citep{DPC-BDFN} employs FKNN kernel functions in local density estimation to enhance inter-cluster separability. 
DPC-FSC \cite{DPC-FSC} innovatively defines a new local density metric via the fuzzy semantic cell model and maximum density principle, achieving a clearer decision graph.
DPC-FWSN \cite{DPC-FWSN} adopts FKNN to balance the contribution of dense and sparse areas and facilitate selection of sparse cluster centers, and proposes an assignment strategy based on the weighted shared neighbor similarity.
Additionally, \citep{AKE-DPC} introduces adaptive KNN to avoid manually selecting hyperparameters, and an evidential assignment strategy to mitigate error propagation

However, these methods face challenges when integrating with deep learning frameworks, due to iterative computations, non-matrix operations, and high computation and memory costs. To mitigate these issues, the proposed DPC-FKNN enables efficient, non-iterative matrix-based computation while maintaining low memory overhead.

\section{Methodology}
\subsection{Overall Architecture}
\begin{figure*}[h] 
  \centering
  \includegraphics[width=0.9\textwidth]{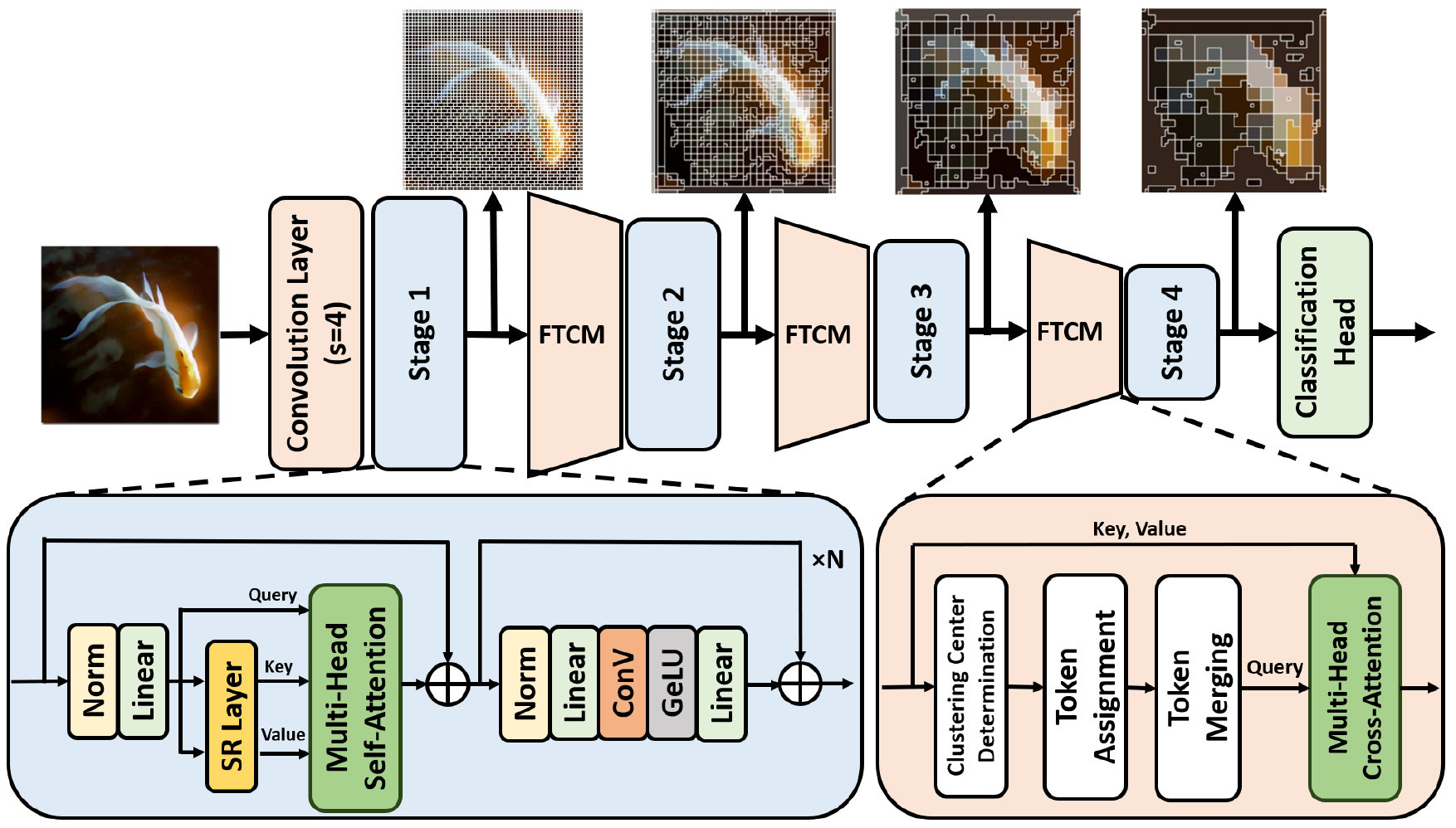} 
  \caption{
  Architecture of the FTCFormer consists of four stages and three FTCM modules. An image is firstly processed by a strided convolutional layer for initial feature extraction. Subsequent stages utilize the FTCM modules for downsampling, each of which performs clustering center determination, token assignment, token merging and token interaction.
  }
  \label{fig:architecture}
\end{figure*}

FTCFormer adopts a similar hierarchical architecture as TCFormer \cite{tcformer}, which comprises four stages with downsampling layers (i.e. the proposed Fuzzy Token Clustering and Merging module) in between, as illustrated in Fig. \ref{fig:architecture}. Each stage consists of a series of basic transformer units, in which the spatial reduction (SR) layer is a strided CNN to reduce the resolution of Key and Value for efficiency. An input image goes through a strided convolutional layer for initial feature extraction and preliminary coarse downsampling. In the subsequent stages, the Fuzzy Token Clustering and Merging (FTCM) module is employed to perform progressive token downsampling and feature refinement based on semantic relevance. The final classification is performed using a linear head that maps the feature maps to image category predictions. The key difference between FTCFormer and TCFormer lies in the design of the FTCM module, which will be elaborated below.

\subsection{Fuzzy Token Clustering and Merging Module}
We propose the \textbf{\underline{F}}uzzy \textbf{\underline{T}}oken \textbf{\underline{C}}lustering and \textbf{\underline{M}}erging (\textbf{\underline{FTCM}}) Module to cluster and merge less important image tokens into flexible, non-fixed tokens. As depicted in Fig. \ref{fig:architecture}, the FTCM module comprises four key stages:
1) Clustering center determination stage computes clustering centers; 
2) Token assignment stage assigns non-clustering-center tokens into clusters; 
3) Token merging stage is to merge all tokens in one cluster into one token; 
4) Token interaction stage utilizes cross-attention mechanism \cite{transformer} to integrate information between original tokens and clustering-based downsampled tokens.
Each stage is detailed in the following subsections.

\subsubsection{Clustering Center Determination}

The first step is to compute representative clustering centers. In TCFormer, the DPC-KNN algorithm \cite{dpcknn} is used for this purpose. However, DPC-KNN is sensitive to the $K$ value \cite{tcformer, dpcknn, IDPC-NNMS} and often yields unstable results when handling outliers and uneven data \cite{DPC-FWSN, IDPC-NNMS} (e.g. learned feature map \cite{ToMe}). 
To address these limitations, we make use of Fuzzy KNN \cite{DPC-FWSN, FDPC, DPC-FSC, FN-DP} and term this clustering method as DPC-FKNN. It considers the influence of the KNN set and FKNN set (tokens outside of KNN), and attempts to reduce the impact of the FKNN set by a distance-based decay. This design balances local reliability with global contextual awareness. By integrating FKNN, DPC-FKNN mitigates the sensitivity issue to the $K$ value, improves resilience to outliers and uneven feature space, and better captures the long-range dependencies inherent in Transformer-based architectures.

To quantify the contribution of each token within FKNN, we propose a fuzzy distance kernel $\mu (i,j)$ between token pairs, which satisfies two essential properties: (1) Distance Monotonicity: closer token pairs receive exponentially higher kernel values; (2) Neighborhood Emphasis: KNN members receive substantially stronger weighting than non-KNN-neighbors. The fuzzy distance kernel $\mu(i, j)$ is formally defined as follows:

\begin{equation}
\mu(i, j) = 
\begin{cases}
\displaystyle \frac{e^{-d_{ij}^2}}{d_{ij} + 1}^2, & \text{if } j \in \text{KNN}(i), \\
\displaystyle \frac{e^{-(\varphi d_{ij})^2}}{(d_{ij} + 1)^2}, & \text{otherwise}.
\end{cases}
\end{equation}
, where KNN(i) represents the K-nearest neighbor set of the token $x_i$; $d_{ij}$ represents the Euclidean distance between token $x_i$ and $x_j$; $\varphi$ is the standard deviation representing the sparsity of all tokens, serving as attenuation for non-KNN-neighbors. Then, we can define the local density $\rho_i$ as follows:

\begin{equation}
\rho_i = \frac{1}{K_{Fuzzy}} \sum_{j=1}^{K_{Fuzzy}} \mu(i,j) + \frac{1}{N} \sum_{j=1}^N \mu(i, j)
\label{eq:density}
\end{equation}
, where K$_{Fuzzy}$ represents the $K$ value in DPC-FKNN. The first term captures the contribution of tokens within the KNN set, emphasizing local structure. While the second term accounts for the influence of all tokens, ensuring global contextual integration.

Then, we compute the distance score $\delta_i$ for token $x_i$. It is defined as the Euclidean distance between each token and the token with the highest local density. As for the token with the highest local density, its distance score is computed as the maximal distance between it and any other tokens.

\begin{equation}
\delta_i = 
\begin{cases} 
\min\limits_{j: \rho_j > \rho_i} \|x_i - x_j\|_2, & \text{if } \exists j \text{ s.t. } \rho_j > \rho_i \\ 
\max\limits_j \|x_i - x_j\|_2, & \text{otherwise}
\end{cases}
\label{eq:delta}
\end{equation}


Cluster centers are selected by maximizing the product $\rho_i \times \delta_i$, with the top $N/4$ scoring tokens (equivalent to 2× downsampling) chosen to maintain hierarchical consistency with standard vision Transformer architectures.

\subsubsection{Token Assignment}
Following the clustering center determination stage, non-center tokens must be assigned to the appropriate clusters. In TCFormer \cite{tcformer}, this assignment is performed using Euclidean distance. However, in high-dimensional or over-parameterized feature spaces, this approach is prone to the curse of dimensionality and sensitive to insignificant noise \cite{ToMe}, which may result in suboptimal token assignment and error propagation to subsequent stages. To overcome this limitation, we propose the Spatial Connectivity Score (SCS) metric, which combines two critical components: (1) the number of shared nearest neighbor (SNN), capturing local topological similarity, and (2) the closeness to their neighbor (CN), reflecting local density and compactness. 

Our assignment strategy prioritizes SCS-based token allocation, using Euclidean distance as a fallback in cases where the SCS score is zero. This hybrid approach ensures robust and context-aware cluster association while preserving spatial coherence within high-dimensional feature embeddings.
SCS is defined as follows:

\begin{equation}
\mathrm{SNN}(i, j) = \{i \in X, j \in X \mid \mathrm{KNN}(i) \cap \mathrm{KNN}(j)\}
\label{eq:snn}
\end{equation}

\begin{equation}
CN(i, j) = \sum_{u \in \mathrm{KNN}(i)} \frac{1}{d_{iu} + 1} + \sum_{\nu \in \mathrm{KNN}(j)} \frac{1}{d_{j\nu} + 1}
\label{eq:proximity}
\end{equation}

\begin{equation}
SCS(i, j) = CN(i,j) \times |SNN(i,j)|
\label{eq:SCS}
\end{equation}
where $|SNN(i,j)|$ is the number of SNN. SCS measures the spatial connection in the high dimension space, which effectively mitigates the issue caused by high-dimensional features; $d_{iu}$ and $d_{jv}$ denote the pairwise Euclidean distances between tokens $(i,u)$ and $(j,v)$, respectively.
Here, we term the K value for the calculation of SCS as $K_{SCS}$.

\subsubsection{Token Merging}
The token merging stage merges tokens within each cluster into a single representative token. The methods in \cite{DynamicVIT} and \cite{tcformer} regress the token-level importance score $P\in \mathbb{R}^{H \times W \times1}$ by a linear layer. However, we argue that employing channel-level importance scores can better preserve the diverse fine-grained semantic information distributed across individual channels. Therefore, we propose the idea of Channel Merging (Cmerge) that enables precise and informative token fusion with the channel-level important score $P\in \mathbb{R}^{H\times W\times C}$. The token merging process can be formulated as follows:

\begin{equation}
    y_{i}^{c} = \frac{\sum_{j \in Cluster_i} e^{P_j^c} x_j^c}{\sum_{j \in Cluster_i} e^{P_j^c}}
\end{equation}
, where $Cluster_i$ represents the i-th cluster;  $x_j^c$ is the $j$-th input token in $c$-the channel; and $P_j^c$ denotes the important score of $j$-th token in $c$-th channel; $y_i^c$ is the merged output for $Cluster_i$ in $c$-th channel.


\subsubsection{Token Interaction}

To enhance the representational robustness of the merged tokens, we employ the cross-attention mechanism that facilitates feature interaction between the original high-resolution tokens and downsampled cluster tokens. This design ensures effective information flow across different resolution scales and helps retain fine-grained details while benefiting from semantic compression. In addition, we incorporate the average of channel-level importance scores into the attention computation, allowing the model to dynamically modulate attention based on the relative significance of token features. The interaction process is formally expressed as:

\begin{equation}
    Attention(Q_m, K_o, V_o) = softmax(\frac{Q_m K_o^T}{\sqrt{d_k}}+avg_c(P)) V
\end{equation}
where $Q_m$ represents the Query coming from the merged tokens; $K_o$ and $V_o$ represent the Key and Value coming from the original token; $avg_c()$ represents average pooling along the C dimension.

\section{Results and Evaluation}

\subsection{Datasets and Implementation Details}
To validate the effectiveness of FTCFormer, we conduct extensive experiments on 32 image classification datasets across diverse domains, as shown in Table~\ref{tab:datasets}. In the absence of officially predefined dataset partitions, we implement a randomized split procedure with an 8:2 ratio to generate training and validation sets.

\begin{table}[htbp]
\renewcommand{\arraystretch}{1.25}
\setlength{\tabcolsep}{2pt}
\resizebox{0.48\textwidth}{!}{ 
\centering
\begin{tabular}{l|l}
\hline
\textbf{Domains} & \textbf{Included Datasets} \\ \hline
Natural Images          & ImageNet-1k \cite{imagenet}, Tiny ImageNet \cite{tinyimagenet}, STL10 \cite{STL10}, \\ 
                        & ImageNette \cite{imagenette}, Caltech101 \cite{caltech101}, Caltech256 \cite{caltech256} \\ \hline
Fine-Grained  & Flowers102 \cite{flowers102}, Food101 \cite{food101}, Standford Cars \cite{car}, \\
                         & Aircraft \cite{aircraft}, Oxford-IIIT Pet \cite{OxfordIIIPet} \\ \hline
Remote Sensing          & WHU-RS19 \cite{WHU-RS19}, RESISC45 \cite{RESISC45}, EuroSAT \cite{EuroSat}, UC Merced \cite{UCMerced} \\ \hline
Medical Image         & Blood Cell \cite{BCCD}, PCAM \cite{PCAM}, SD-198 \cite{SD-198} \\ \hline
MNIST-like             & MNIST \cite{MNIST}, FMNIST, SVHN, QMNIST \\
                        & EMNIST$_{\text{letter}}$, EMNIST$_{\text{byclass}}$ , KMNIST \\ \hline
CIFAR-like             & CIFAR-10 \cite{CIFAR}, CIFAR-100 \cite{CIFAR}, GTSRB, CINIC-10 \\ \hline
Other Domains           & DTD \cite{DTD}, ImageNet-Sketch \cite{imagentsketch}, FER2013 \cite{FER2013} \\ \hline
\end{tabular}}
\caption{Datasets used in the experiments, organized into seven domains.}
\label{tab:datasets}
\end{table}

Our experimental setup follows PVT~\cite{PVT} with various data augmentation methods including random cropping, random horizontal flipping \cite{41}, label smoothing \cite{42}, Mixup \cite{43}, CutMix \cite{44} and random erasing \cite{45}. We employ the AdamW optimizer \cite{46} with momentum of 0.9 and weight decay of 0.05, cosine learning rate schedule \cite{47} and a 5-epoch linear warm-up at the beginning of training. Hyperparameters $K_{Fuzzy}$ and $K_{SCS}$ are set to 5 for all datasets. The model contains 2 Transformer blocks in each stage. For datasets containing large-size images (> $64 \times 64$ resolution), we resize them to $224 \times 224$, while for small-size images we resize them to $64 \times 64$. Specific training hyperparameters, including the number of epochs, learning rate, batch size and training devices configuration may vary across datasets, with complete details available in our GitHub repository's log files. Notably, for ImageNet-1k \cite{imagenet} we train the model using three A100 GPUs with the learning rate 0.001, batch size 120 for 300 epochs. 

\subsection{Overall Results}

\begin{figure*}[h] 
  \centering
  \includegraphics[width=1.0\textwidth]{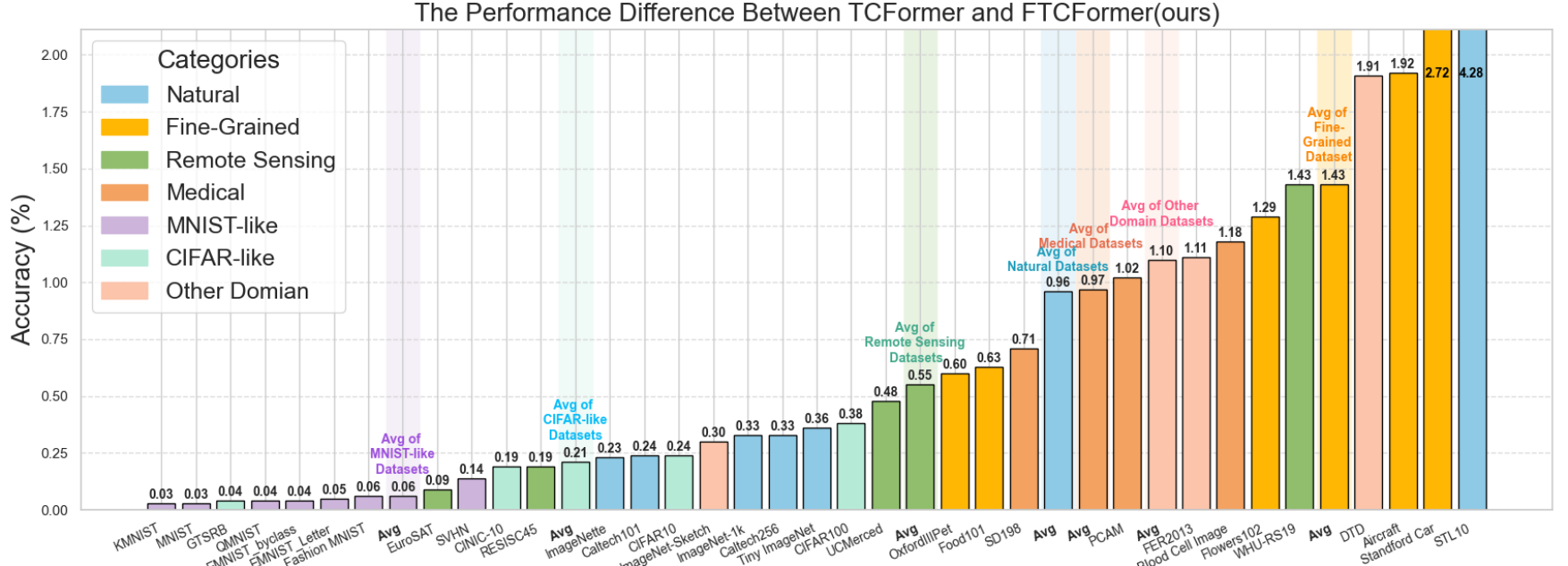} 
  \caption{Performance difference in terms of classification accuracy between FTCFormer (ours) and TCFormer (baseline) on 32 datasets across various domains. Average improvement for each domain is also shown.}
  \label{fig:difference}
\end{figure*}

We evaluate the performance of FTCFormer on 32 datasets spanning various domains and compare the results with TCFormer in Fig.~\ref{fig:difference}. Overall, FTCFormer consistently outperforms TCFormer across all datasets in various domains, demonstrating its strong generalization ability and robustness. Specifically, the average performance improvements over the baseline are 1.43\%, 1.09\%, 0.55\%, 0.97\%, 0.21\%, and 0.06\% on fine-grained, natural, remote sensing, medical, MNIST-like, and CIFAR-like datasets, respectively. The observed performance improvements demonstrate that the FTCM module can effectively adapt across diverse visual domains and exhibits strong generalizability.

\begin{table*}[htbp]
\centering
\resizebox{\textwidth}{!}{
\begin{tabular}{ccc}  
\begin{minipage}{0.53\textwidth}
\centering
\resizebox{\textwidth}{!}{
\begin{tabular}{c|c|c|c}
\toprule
\textbf{Method} & \textbf{Flowers102} & \textbf{Method} & \textbf{Stanford Cars} \\
\hline
CVT \cite{CVT}               & 56.29 & gMLP-Ti/16 \cite{gMLP}   & 78.70 \\ 
ResNet50 \cite{ResNet}       & 63.20 & RDGC \cite{RDGC}         & 80.30  \\ 
MobileNet \cite{Mobilenets}  & 70.06  & ResNet50 \cite{ResNet}   & 82.50 \\ 
F-SAM \cite{fsam}            & 75.15 & ViT-M/16 \cite{ViT}      & 83.89 \\ 
TCFormer \cite{tcformer}     & 77.83 & TCFormer \cite{tcformer} & 81.83 \\ 
\rowcolor{gray!20} FTCFormer & 79.12 & FTCFormer                & 84.55 \\
\bottomrule
\end{tabular}
}
\subcaption{Benchmark on fine-grained datasets: Flowers102 \cite{flowers102} and Stanford Car \cite{car}.}
\label{tab:sub1}
\end{minipage}
&
\begin{minipage}{0.24\textwidth}
\centering
\resizebox{\textwidth}{!}{
\begin{tabular}{c|c}
\toprule
\textbf{Method} & \textbf{RESUS45}   \\
\hline
StarNet \cite{StarNet}   & 93.08  \\ 
EffiFormer \cite{EffienctVFormer} & 95.14 \\ 
ResNet50 \cite{ResNet}   & 95.65 \\ 
LWGANet \cite{LWGANet}   & 96.17  \\ 
TCFormer \cite{tcformer} & 96.43  \\ 
\rowcolor{gray!20} FTCFormer  & 96.62 \\
\bottomrule
\end{tabular}
}
\subcaption{Benchmark on remote sensing dataset: RESISC45 \cite{RESISC45}.}
\label{tab:sub2}
\end{minipage}
&
\begin{minipage}{0.23\textwidth}
\centering
\resizebox{\textwidth}{!}{
\begin{tabular}{c|c}
\toprule
\textbf{Method} & \textbf{PCAM}   \\
\hline

DARC \cite{DARC}                & 83.01  \\ 
ViT-L  \cite{ViT}               & 86.72 \\
ResNet50 \cite{ResNet}          & 86.73  \\ 
MaskedFeat \cite{MaskedFeat}    & 87.00 \\ 
TCFormer \cite{tcformer}        & 86.47  \\ 
\rowcolor{gray!20} FTCFormer    & 87.49 \\
\bottomrule
\end{tabular}
}
\subcaption{Benchmark on medical dataset: PCAM \cite{PCAM}.}
\label{tab:sub3}
\end{minipage}

\end{tabular}
}
\vspace{0.2cm}
\caption{Performance comparison of FTCFormer against leading methods on representative datasets of three domains on classification accuracy.}
\label{tab:main_table}
\end{table*}

Notably, the most substantial performance gain is observed on \textbf{fine-grained datasets}, where FTCFormer achieves an average accuracy improvement of 1.43\%. This indicates that the FTCM module enhances the ability to discern subtle feature differences through the refined clustering and merging stage, leading to more precise feature aggregation. Similar improvement is observed on \textbf{natural datasets}, with an average increase of 1.09\%. These confirm the effectiveness of the FTCM module in generating more discriminative and semantically coherent token representations.

For \textbf{medical imaging datasets}, FTCM yields an average accuracy improvement of 0.97\%, demonstrating its adaptability to medical images characterized by structured anatomical patterns, complex tissue organization, and subtle pathological variations. 
Similarly, on \textbf{remote sensing datasets}, FTCM achieves 0.55\% average accuracy improvement, confirming its robustness in processing high-resolution aerial imagery with extensive spatial coverage and complex backgrounds.

In simple and low-resolution \textbf{MNIST-like and CIFAR-like datasets}, FTCM only produces marginal performance improvements of 0.21\% and 0.06\%, respectively. Nevertheless, this demonstrates its consistent effectiveness in near-saturation scenarios, considering that performance on some datasets exceeds 99\% accuracy. On datasets in \textbf{other domains}, FTCFormer still outperforms the baseline on sketch, facial emotion and texture classification.

\subsection{Comparison with Leading methods}

We establish benchmarks on natural image (ImageNet-1k \cite{imagenet} in Table \ref{tab:imagenet1k}), fine-grained (Flowers102~\cite{flowers102} and Stanford Cars~\cite{car} in Table~\ref{tab:sub1}), remote sensing (RESISC45~\cite{RESISC45} in Table~\ref{tab:sub2}), and medical imaging domains (PCAM~\cite{PCAM} in Table~\ref{tab:sub3}). These results confirm that the proposed FTCM module exhibits strong generalization capabilities across diverse visual domains, effectively handling tasks ranging from general object recognition to specialized downstream classification tasks. The consistent improvement highlights the robustness of FTCM in various classification scenarios.

\begin{table}[h]
\centering
\begin{tabular}{lccc}
\toprule
Method                              & \#Param. (M) & FLOPs (G) & Top-1 Acc. (\%) \\
\midrule
PVT \cite{PVT}                       & 13.2 & 1.9 & 75.1 \\
ResNet50 \cite{ResNet}               & 25.6 & 4.1 & 76.1 \\
ViT-L/16 \cite{ViT}                  & 307  & 190 & 76.5 \\
CI2P-ViT \cite{CI2PViT}              & 89.0 & 8,5 & 77.0 \\
StarNet-S3 \cite{StarNet}            & 5.8  & 0.8 & 77.3 \\ 
ResNet101\cite{ResNet}               & 44.7 & 7.9 & 77.4 \\
TCFormer \cite{tcformer}             & 14.1 & 3.8 & 77.5 \\
ResNeXt50 \cite{ResNeXt}             & 25.0 & 4.3 & 77.6 \\
\rowcolor{gray!20} FTCFormer (Ours)  & 14.6 & 4.1 & 77.9 \\
\bottomrule
\end{tabular}
\caption{Benchmark on natural image dataset: ImageNet-1k \cite{imagenet}.}
\label{tab:imagenet1k}
\end{table}

\subsection{Qualitative Evaluation}
\begin{figure*}[t!] 
  \centering
  \includegraphics[width=\textwidth]{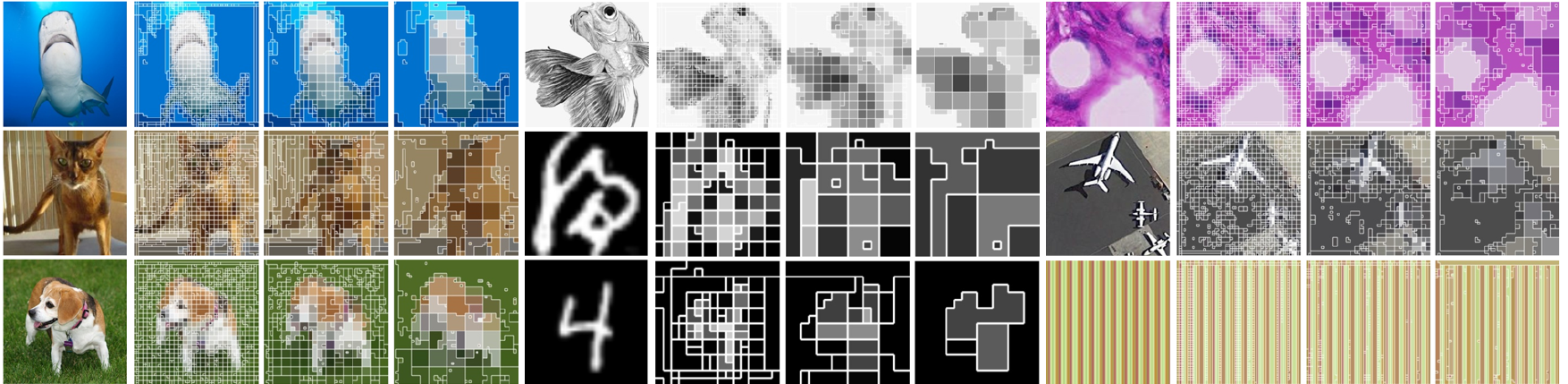} 
  \caption{Visualization of FTCFormer tokenization results across different datasets.
  , including natural, fine-grained, medical imagery, remote sensing and MNIST-like datasets. Key areas, such as faces, digits, lesions and structural components, are assigned more and finer tokens, preserving detailed structures, while less informative image regions, such as background, are represented by fewer and coarser tokens. }
  \vspace{0.1cm}
  \label{fig:vis}
\end{figure*}

As illustrated in Fig.~\ref{fig:vis}, the proposed method produces meaningful tokenization results across various domains. It can be seen that semantically important regions, such as faces, digits, lesions and structural components, are assigned a greater number of finer-grained vision tokens. Conversely, less important areas, particularly background regions, are represented by coarser and fewer tokens.
This observation aligns with our design objective, which aims to dynamically generate vision tokens based on the semantic meanings, enabling more precise encoding of critical information. Such a strategy leads to not only improved performance on image classification, but also enhanced interpretability of classification results.

\begin{table}[H]
    \centering
    \renewcommand{\arraystretch}{1.15}  
    \resizebox{\linewidth}{!}{ 
    \begin{tabular}{c|c|c|c} \toprule
        \makecell{Downsampling\\Methods} & Flower102 & FER2012 & CIFAR10 \\\hline
        \makecell{Conv$_{\text{k22s3}}$} & 76.92\textcolor{blue}{(-1.42)} & 69.01\textcolor{blue}{(-1.08)} & 92.74\textcolor{blue}{(-1.34)} \\
        \makecell{Conv$_{\text{k33s2}}$} & 77.70\textcolor{blue}{(-2.20)} & 69.55\textcolor{blue}{(-1.62)} & 92.84\textcolor{blue}{(-1.44)} \\
        MaxPooling & 78.89\textcolor{blue}{(-0.23)} & 69.88\textcolor{blue}{(-0.75)} & 92.82\textcolor{blue}{(-1.36)} \\
        AvgPooling & 77.74\textcolor{blue}{(-1.38)} & 68.51\textcolor{blue}{(-2.12)} & 92.82\textcolor{blue}{(-1.36)} \\
        TCM & 77.83\textcolor{blue}{(-1.29)} & 69.52\textcolor{blue}{(-1.11)} & 93.94\textcolor{blue}{(-0.24)}\\
        \rowcolor{gray!20} FTCM & 79.12 & 70.63 & 94.18 \\ \bottomrule
    \end{tabular}
    }
    \caption{Performance comparison of clustering-based and grid-based downsampling methods on Flowers102 \cite{flowers102}, FER2013 \cite{FER2013} and CIFAR-10 \cite{CIFAR} datasets. Results show Top-1 accuracy (\%) with performance degradation relative to our proposed FTCM method in blue.
    }
    \label{tab:grid-based}
\end{table}

\subsection{Ablation Study}
This section provides a detailed ablation study on the proposed FTCM (as a downsampling module), each component in FTCM (i.e. DPC-FKNN, SCS and Cmerge), and hyperparameters $K_{Fuzzy}$ and $K_{SCS}$.

\begin{table*}[htbp]
\centering
\resizebox{\linewidth}{!}{ 
\begin{tabular}{ccc|llll|ll}
\toprule
\multicolumn{3}{c|}{Components} & \multicolumn{4}{c|}{Datasets} & & \\ \hline
DPC-FKNN & SCS & Cmerge & \multicolumn{1}{c}{Flowers102} & \multicolumn{1}{c}{DTD} & \multicolumn{1}{c}{Stanford Cars} & \multicolumn{1}{c|}{Blood Cell Image} & GFLOPs & Parameter (M) \\ \hline
 & & & 77.83 & 52.66 & 81.83 & 88.73 & 3.84 & 14.23 \\
\Checkmark & & & 78.63 \textcolor{blue}{(+0.80)} & 53.62 \textcolor{blue}{(+0.96)} & 84.72 \textcolor{blue}{(+2.89)} & 89.10 \textcolor{blue}{(+0.37)} & 3.86 \textcolor{red}{(+0.02)} & 14.23 (+0.00) \\
\Checkmark & \Checkmark & & 78.79 \textcolor{blue}{(+0.16)} & 53.99 \textcolor{blue}{(+0.37)} & 85.33 \textcolor{blue}{(+0.61)} & 89.18 \textcolor{blue}{(+0.08)} & 3.91 \textcolor{red}{(+0.05)} & 14.23 (+0.00) \\
\Checkmark & \Checkmark & \Checkmark & 79.12 \textcolor{blue}{(+0.33)} & 54.20 \textcolor{blue}{(+0.21)} & 84.55 \textcolor{blue}{(+0.22)} & 89.91 \textcolor{blue}{(+0.73)} & 4.10 \textcolor{red}{(+0.19)} & 14.61 \textcolor{red}{(+0.38)} \\ \hline
\multicolumn{3}{c|}{Total Improvement} & {\color{blue} +1.29} & {\color{blue} +1.55} & {\color{blue} +2.72} & {\color{blue} +1.18} & {\color{red} +0.26} & {\color{red} +0.38} \\ \bottomrule
\end{tabular}
} 
\caption{Ablation study of proposed model components (DPC-FKNN, SCS, and Cmerge) on Flowers102 \cite{flowers102}, DTD \cite{DTD}, Stanford Cars \cite{car} and Blood Cell Image \cite{BCCD}. Performance metrics show Top-1 accuracy (\%) with relative improvement in blue. An increase in computation is in red. The bottom row summarizes total improvements when incorporating all components.}
\label{tab:components}
\end{table*}

\begin{figure*}[t!] 
    \centering
    \begin{subfigure}[b]{0.32\textwidth}
        \includegraphics[width=\linewidth]{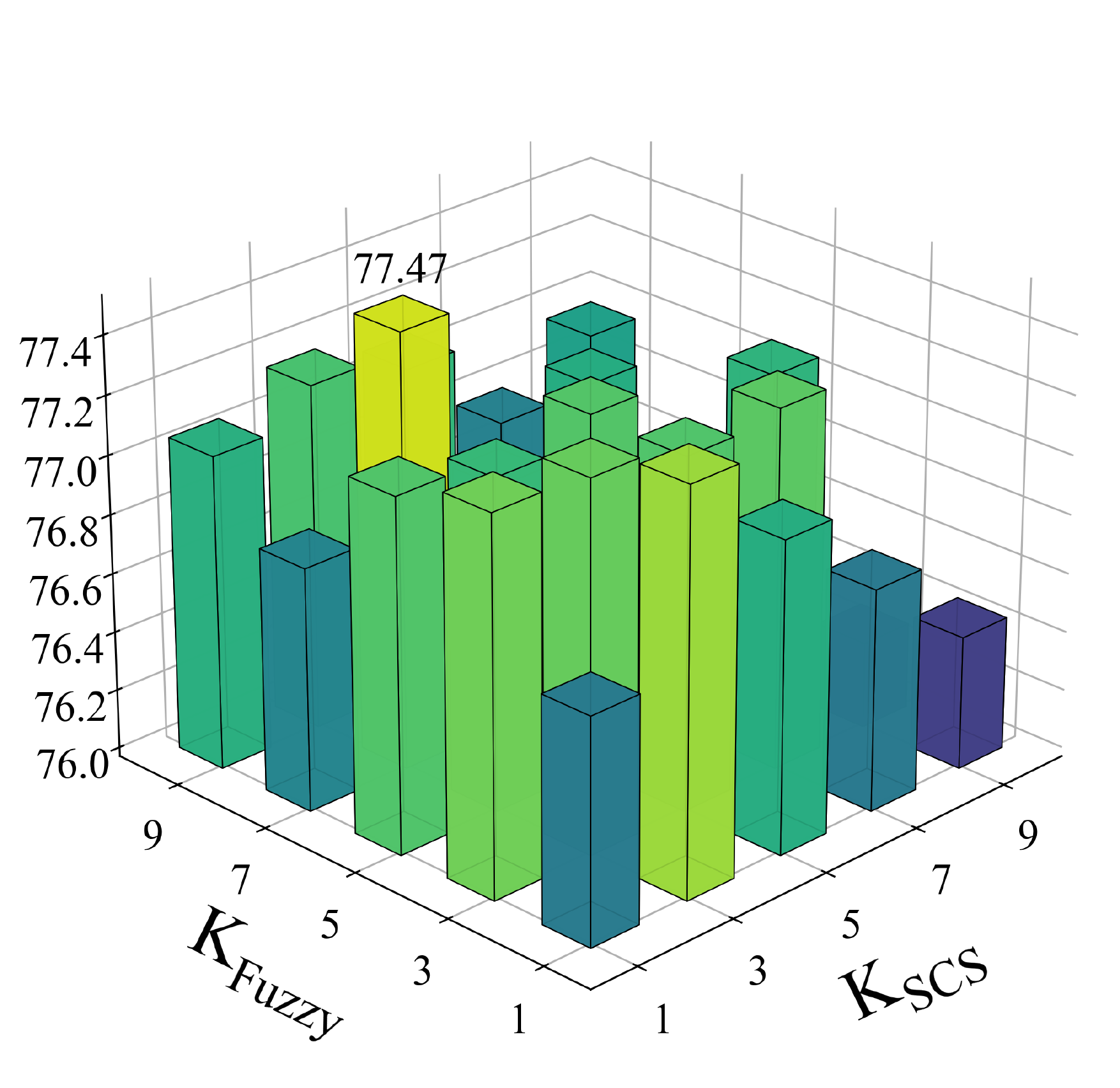}
        \caption{On CIFAR100 \cite{CIFAR}.}
        \vspace{0.5cm}
        \label{fig:cifar100}
    \end{subfigure}
    \hfill
    \begin{subfigure}[b]{0.32\textwidth}
        \includegraphics[width=\linewidth]{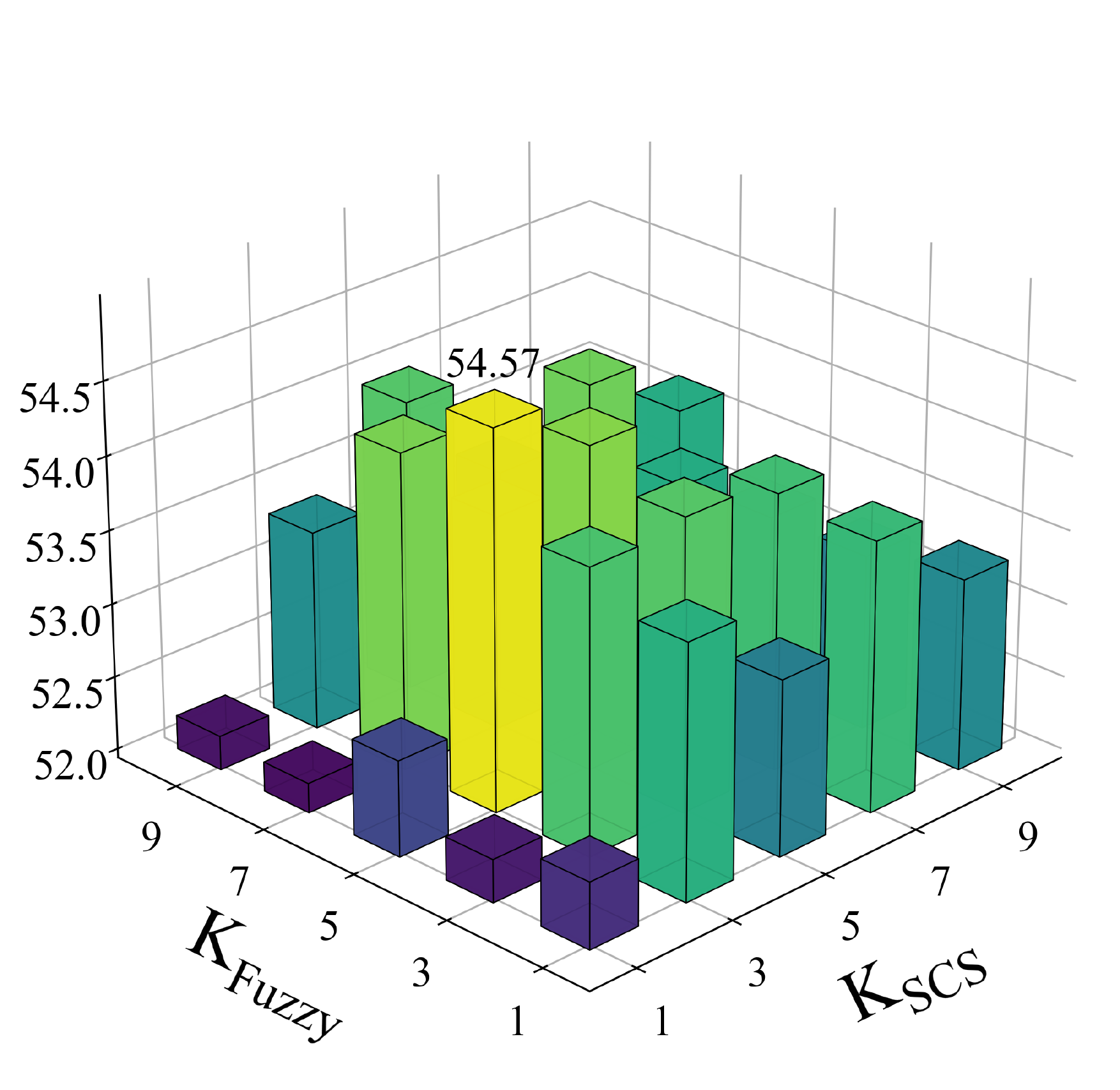}
        \caption{On DTD \cite{DTD}.}
        \vspace{0.5cm}
        \label{fig:dtd}
    \end{subfigure}
    \hfill
    \begin{subfigure}[b]{0.32\textwidth}
        \includegraphics[width=\linewidth]{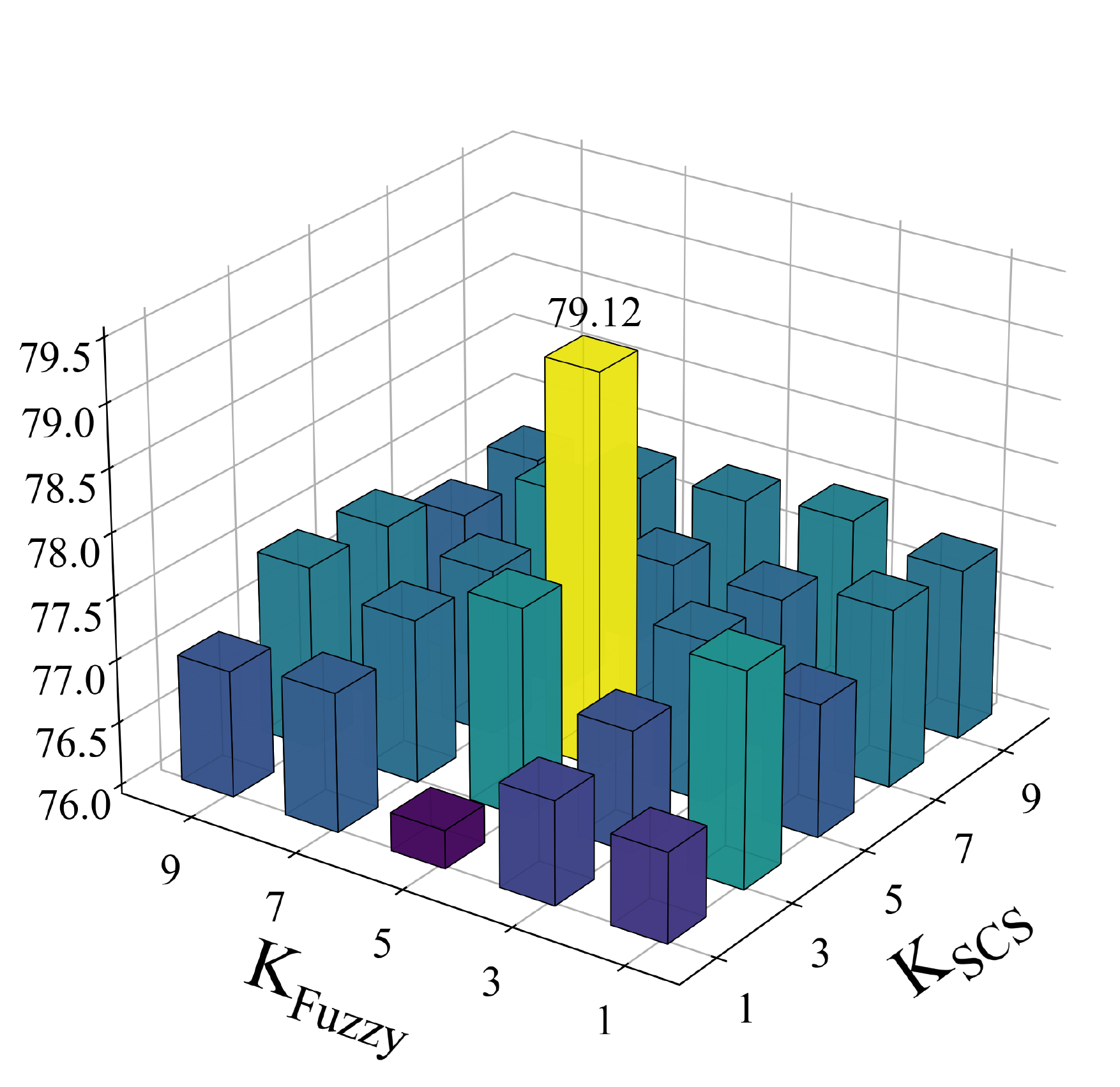}
        \caption{On Flowers102 \cite{flowers102}.}
        \vspace{0.5cm}
        \label{fig:flowers102}
    \end{subfigure}
    \caption{Hyperparameter ablation study examining the impact of $K_{Fuzzy}$ (for DPC-FKNN) and $K_{SCS}$ (for SCS metric) across three datasets. Performance is evaluated using Top-1 accuracy with $K$ values ranging from 1 to 9 and an interval of 2. Optimal settings are explicitly annotated with their corresponding accuracy scores.}
    \label{fig:hyperparameter}
\end{figure*}

\subsubsection{Effect of FTCM}
Table~\ref{tab:grid-based} illustrates the comparative performance between grid-based and clustering-based downsampling methods in terms of Top-1 accuracy. FTCM consistently outperforms FTM and all conventional grid-based downsampling methods across three datasets. Notably, the baseline TCM does not outperform MaxPooling on both Flowers102~\cite{flowers102} and FER2013~\cite{FER2013} datasets by 1.06\% and 0.36\%, respectively. This further highlights the benefits and effectiveness of FTCM in feature representation learning for image classification tasks.

\subsubsection{Effect of Individual Components in FTCM}
As demonstrated in Table~\ref{tab:components}, we evaluate the contributions of each individual component in FTCM across four benchmark datasets. \textbf{DPC-FKNN} emerges as the most beneficial module, delivering substantial performance improvement; most notably, 2.89\% improvement on the Stanford Cars dataset \cite{car}. This significant enhancement validates the idea that Fuzzy-KNN can effectively mitigate the inherent limitations (e.g. inadaptability to learned feature maps) of the baseline method DPC-KNN, by determining more accurate clustering centers.
\textbf{SCS} helps achieve auxiliary while consistent improvements on all four datasets, particularly, a 0.61\% improvement on the Stanford Cars dataset \cite{car} is observed. These improvements confirm the importance of incorporating spatial connectivity metrics beyond Euclidean distance during token assignment, especially in high-dimensional feature space.
\textbf{Cmerge} further enhances performance, underscoring the necessity of channel-wise importance scores for token merging. This also underscores that different channels of a token should be given different important scores to merge.
As for \textbf{computation}, the entire FTCM introduces only marginal overhead (+0.26 GFLOPs and +0.38M parameters), maintaining acceptable computation while improving the overall performance. This demonstrates a favorable trade-off between model complexity and performance improvements.

\subsubsection{Effect of Hyperparameters $K_{Fuzzy}$ and $K_{SCS}$}


We also evaluate the impact of two critical hyperparameters: $K_{Fuzzy}$ (for DPC-FKNN clustering) and $K_{SCS}$ (for SCS metric computation) on CIFAR100 \cite{CIFAR}, DTD \cite{DTD} and Flowers102 \cite{flowers102}. The results are plotted in Fig.~\ref{fig:hyperparameter}.
In general, it can be observed that the optimal performance consistently occurs near (5,5) across three datasets, though minor variations may exist because of specific characteristics of different domains.
K$_{Fuzzy}$ exhibits a certain degree of robustness except K$_{SCS}$=1, validating one of the design objectives to reduce hyperparameter sensitivity. 
For K$_{SCS}$, the figure reveals significant performance degradation when setting it to 1, which is equivalent to adopting Euclidean distance. This finding undoubtedly verifies the necessity and effectiveness of taking into account spatial connectivity. A smaller K$_{SCS}$ restricts the spatial connectivity evaluation scope, while a larger one may potentially introduce noise from distant tokens. The best performance around K$_{SCS}$=5 suggests an optimal balance between preserving locality and incorporating global context.

\section{Conclusion}
We propose \textbf{\underline{F}}uzzy \textbf{\underline{T}}oken \textbf{\underline{C}}lustering Trans\textbf{\underline{former}} (\textbf{FTCFormer}), which incorporates a novel clustering-based downsampling module, \textbf{FTCM}, to dynamically generate vision tokens based on semantic contents inside images. It can allocate fewer tokens to less informative regions while directing attention on semantically meaningful areas, regardless of spatial adjacency or irregular shapes. FTCM employs Density Peak Clustering with Fuzzy K-Nearest Neighbors (DPC-FKNN) for robust clustering center determination, a spatial connectivity-aware similarity metric for token assignment, and a channel-wise merging strategy to preserve fine-grained semantic features during token fusion. Extensive experiments across 32 datasets demonstrate that FTCFormer consistently outperforms other token aggregation methods in image classification tasks, validating its effectiveness and generalization ability across a wide range of visual domains. Future work will explore strategies to accelerate training on high-resolution images and extend FTCFormer's clustering principles to convolutional neural networks.

\bibliography{mybibfile}

\begin{thebibliography}{73}
\providecommand{\natexlab}[1]{#1}
\providecommand{\url}[1]{\texttt{#1}}
\expandafter\ifx\csname urlstyle\endcsname\relax
  \providecommand{\doi}[1]{doi: #1}\else
  \providecommand{\doi}{doi: \begingroup \urlstyle{rm}\Url}\fi

\bibitem[Bian et~al.(2020)Bian, Chung, and Wang]{FDPC}
Z.~Bian, F.-L. Chung, and S.~Wang.
\newblock Fuzzy density peaks clustering.
\newblock \emph{IEEE Transactions on Fuzzy Systems}, 29\penalty0 (7):\penalty0 1725--1738, 2020.

\bibitem[Bolya et~al.(2022)Bolya, Fu, Dai, Zhang, Feichtenhofer, and Hoffman]{ToMe}
D.~Bolya, C.-Y. Fu, X.~Dai, P.~Zhang, C.~Feichtenhofer, and J.~Hoffman.
\newblock Token merging: Your vit but faster.
\newblock \emph{arXiv preprint arXiv:2210.09461}, 2022.

\bibitem[Bossard et~al.(2014)Bossard, Guillaumin, and Van~Gool]{food101}
L.~Bossard, M.~Guillaumin, and L.~Van~Gool.
\newblock Food-101 -- mining discriminative components with random forests.
\newblock In \emph{European Conference on Computer Vision}, 2014.

\bibitem[Carion et~al.(2020)Carion, Massa, Synnaeve, Usunier, Kirillov, and Zagoruyko]{DETR}
N.~Carion, F.~Massa, G.~Synnaeve, N.~Usunier, A.~Kirillov, and S.~Zagoruyko.
\newblock End-to-end object detection with transformers.
\newblock In \emph{European conference on computer vision}, pages 213--229. Springer, 2020.

\bibitem[Chen et~al.(2021)Chen, Wang, Zhang, Huo, and Gao]{RDGC}
W.~Chen, C.~Wang, Z.~Zhang, Z.~Huo, and L.~Gao.
\newblock Reweighted dynamic group convolution.
\newblock In \emph{ICASSP 2021-2021 IEEE International Conference on Acoustics, Speech and Signal Processing (ICASSP)}, pages 3940--3944. IEEE, 2021.

\bibitem[Cheng et~al.(2017)Cheng, Han, and Lu]{RESISC45}
G.~Cheng, J.~Han, and X.~Lu.
\newblock Remote sensing image scene classification: Benchmark and state of the art.
\newblock \emph{Proceedings of the IEEE}, 105\penalty0 (10):\penalty0 1865--1883, 2017.

\bibitem[Cimpoi et~al.(2014)Cimpoi, Maji, Kokkinos, Mohamed, , and Vedaldi]{DTD}
M.~Cimpoi, S.~Maji, I.~Kokkinos, S.~Mohamed, , and A.~Vedaldi.
\newblock Describing textures in the wild.
\newblock In \emph{Proceedings of the {IEEE} Conf. on Computer Vision and Pattern Recognition ({CVPR})}, 2014.

\bibitem[Coates et~al.(2011)Coates, Ng, and Lee]{STL10}
A.~Coates, A.~Ng, and H.~Lee.
\newblock An analysis of single-layer networks in unsupervised feature learning.
\newblock In \emph{Proceedings of the fourteenth international conference on artificial intelligence and statistics}, pages 215--223. JMLR Workshop and Conference Proceedings, 2011.

\bibitem[Dai and Yang(2010)]{WHU-RS19}
D.~Dai and W.~Yang.
\newblock Satellite image classification via two-layer sparse coding with biased image representation.
\newblock \emph{IEEE Geoscience and remote sensing letters}, 8\penalty0 (1):\penalty0 173--176, 2010.

\bibitem[Deng et~al.(2009)Deng, Dong, Socher, Li, Li, and Fei-Fei]{imagenet}
J.~Deng, W.~Dong, R.~Socher, L.-J. Li, K.~Li, and L.~Fei-Fei.
\newblock Imagenet: A large-scale hierarchical image database.
\newblock In \emph{2009 IEEE conference on computer vision and pattern recognition}, pages 248--255. Ieee, 2009.

\bibitem[Ding et~al.(2023{\natexlab{a}})Ding, Du, Xu, Shi, Wang, and Li]{IDPC-NNMS}
S.~Ding, W.~Du, X.~Xu, T.~Shi, Y.~Wang, and C.~Li.
\newblock An improved density peaks clustering algorithm based on natural neighbor with a merging strategy.
\newblock \emph{Information Sciences}, 624:\penalty0 252--276, 2023{\natexlab{a}}.

\bibitem[Ding et~al.(2023{\natexlab{b}})Ding, Du, Xu, Shi, Wang, and Li]{SFKNN-DPC}
S.~Ding, W.~Du, X.~Xu, T.~Shi, Y.~Wang, and C.~Li.
\newblock An improved density peaks clustering algorithm based on natural neighbor with a merging strategy.
\newblock \emph{Information Sciences}, 624:\penalty0 252--276, 2023{\natexlab{b}}.

\bibitem[Dosovitskiy et~al.(2020)Dosovitskiy, Beyer, Kolesnikov, Weissenborn, Zhai, Unterthiner, Dehghani, Minderer, Heigold, Gelly, et~al.]{ViT}
A.~Dosovitskiy, L.~Beyer, A.~Kolesnikov, D.~Weissenborn, X.~Zhai, T.~Unterthiner, M.~Dehghani, M.~Minderer, G.~Heigold, S.~Gelly, et~al.
\newblock An image is worth 16x16 words: Transformers for image recognition at scale.
\newblock \emph{arXiv preprint arXiv:2010.11929}, 2020.

\bibitem[Du et~al.(2016)Du, Ding, and Jia]{dpcknn}
M.~Du, S.~Ding, and H.~Jia.
\newblock Study on density peaks clustering based on k-nearest neighbors and principal component analysis.
\newblock \emph{Knowledge-Based Systems}, 99:\penalty0 135--145, 2016.

\bibitem[Du et~al.(2018)Du, Ding, and Xue]{FN-DP}
M.~Du, S.~Ding, and Y.~Xue.
\newblock A robust density peaks clustering algorithm using fuzzy neighborhood.
\newblock \emph{International Journal of Machine Learning and Cybernetics}, 9:\penalty0 1131--1140, 2018.

\bibitem[Fayyaz et~al.(2022)Fayyaz, Koohpayegani, Jafari, Sengupta, Joze, Sommerlade, Pirsiavash, and Gall]{ATS}
M.~Fayyaz, S.~A. Koohpayegani, F.~R. Jafari, S.~Sengupta, H.~R.~V. Joze, E.~Sommerlade, H.~Pirsiavash, and J.~Gall.
\newblock Adaptive token sampling for efficient vision transformers.
\newblock In \emph{European Conference on Computer Vision}, pages 396--414. Springer, 2022.

\bibitem[Fei-Fei et~al.(2006)Fei-Fei, Fergus, and Perona]{caltech101}
L.~Fei-Fei, R.~Fergus, and P.~Perona.
\newblock One-shot learning of object categories.
\newblock \emph{IEEE transactions on pattern analysis and machine intelligence}, 28\penalty0 (4):\penalty0 594--611, 2006.

\bibitem[Goodfellow et~al.(2013)Goodfellow, Erhan, Carrier, Courville, Mirza, Hamner, Cukierski, Tang, Thaler, Lee, et~al.]{FER2013}
I.~J. Goodfellow, D.~Erhan, P.~L. Carrier, A.~Courville, M.~Mirza, B.~Hamner, W.~Cukierski, Y.~Tang, D.~Thaler, D.-H. Lee, et~al.
\newblock Challenges in representation learning: A report on three machine learning contests.
\newblock In \emph{Neural information processing: 20th international conference, ICONIP 2013, daegu, korea, november 3-7, 2013. Proceedings, Part III 20}, pages 117--124. Springer, 2013.

\bibitem[Griffin et~al.(2007)Griffin, Holub, Perona, et~al.]{caltech256}
G.~Griffin, A.~Holub, P.~Perona, et~al.
\newblock Caltech-256 object category dataset.
\newblock Technical report, Technical Report 7694, California Institute of Technology Pasadena, 2007.

\bibitem[He et~al.(2016)He, Zhang, Ren, and Sun]{ResNet}
K.~He, X.~Zhang, S.~Ren, and J.~Sun.
\newblock Deep residual learning for image recognition.
\newblock In \emph{Proceedings of the IEEE conference on computer vision and pattern recognition}, pages 770--778, 2016.

\bibitem[Helber et~al.(2019)Helber, Bischke, Dengel, and Borth]{EuroSat}
P.~Helber, B.~Bischke, A.~Dengel, and D.~Borth.
\newblock Eurosat: A novel dataset and deep learning benchmark for land use and land cover classification.
\newblock \emph{IEEE Journal of Selected Topics in Applied Earth Observations and Remote Sensing}, 12\penalty0 (7):\penalty0 2217--2226, 2019.

\bibitem[Howard(2017)]{Mobilenets}
A.~G. Howard.
\newblock Mobilenets: Efficient convolutional neural networks for mobile vision applications.
\newblock \emph{arXiv preprint arXiv:1704.04861}, 2017.

\bibitem[Howard(2019)]{imagenette}
J.~Howard.
\newblock Imagenette: A smaller subset of imagenet for quick experiments, 2019.
\newblock URL \url{https://github.com/fastai/imagenette}.
\newblock Accessed: 2025-04-09.

\bibitem[Ji et~al.(2022)Ji, Li, Zhang, Zhang, Yang, Yin, and Wang]{AKE-DPC}
F.~Ji, L.~Li, T.~Zhang, B.~Zhang, J.~Yang, J.~Yin, and Q.~Wang.
\newblock A density peak clustering algorithm based on adaptive k-nearest neighbors with evidential strategy.
\newblock In \emph{Proceedings of the 2022 6th International Conference on Computer Science and Artificial Intelligence}, pages 166--171, 2022.

\bibitem[Kim et~al.(2024)Kim, Gao, Hsu, Shen, and Jin]{ToFu}
M.~Kim, S.~Gao, Y.-C. Hsu, Y.~Shen, and H.~Jin.
\newblock Token fusion: Bridging the gap between token pruning and token merging.
\newblock In \emph{Proceedings of the IEEE/CVF Winter Conference on Applications of Computer Vision}, pages 1383--1392, 2024.

\bibitem[Kong et~al.(2022)Kong, Dong, Ma, Meng, Niu, Sun, Shen, Yuan, Ren, Tang, et~al.]{SPViT}
Z.~Kong, P.~Dong, X.~Ma, X.~Meng, W.~Niu, M.~Sun, X.~Shen, G.~Yuan, B.~Ren, H.~Tang, et~al.
\newblock Spvit: Enabling faster vision transformers via latency-aware soft token pruning.
\newblock In \emph{European conference on computer vision}, pages 620--640. Springer, 2022.

\bibitem[Krause et~al.(2013)Krause, Stark, Deng, and Fei-Fei]{car}
J.~Krause, M.~Stark, J.~Deng, and L.~Fei-Fei.
\newblock 3d object representations for fine-grained categorization.
\newblock In \emph{Proceedings of the IEEE international conference on computer vision workshops}, pages 554--561, 2013.

\bibitem[Krizhevsky et~al.(2009)Krizhevsky, Hinton, et~al.]{CIFAR}
A.~Krizhevsky, G.~Hinton, et~al.
\newblock Learning multiple layers of features from tiny images.
\newblock 2009.

\bibitem[Le and Yang(2015)]{tinyimagenet}
Y.~Le and X.~Yang.
\newblock Tiny imagenet visual recognition challenge.
\newblock \emph{CS 231N}, 7\penalty0 (7):\penalty0 3, 2015.

\bibitem[LeCun et~al.(1998)LeCun, Bottou, Bengio, and Haffner]{MNIST}
Y.~LeCun, L.~Bottou, Y.~Bengio, and P.~Haffner.
\newblock Gradient-based learning applied to document recognition.
\newblock \emph{Proceedings of the IEEE}, 86\penalty0 (11):\penalty0 2278--2324, 1998.

\bibitem[Lee et~al.(2024)Lee, Choi, and Kim]{MCTF}
S.~Lee, J.~Choi, and H.~J. Kim.
\newblock Multi-criteria token fusion with one-step-ahead attention for efficient vision transformers.
\newblock In \emph{Proceedings of the IEEE/CVF Conference on Computer Vision and Pattern Recognition}, pages 15741--15750, 2024.

\bibitem[Li et~al.(2022{\natexlab{a}})Li, Liu, Yue, Cheng, Kuang, Bai, Wang, and Wang]{DARC}
J.~Li, J.~Liu, H.~Yue, J.~Cheng, H.~Kuang, H.~Bai, Y.~Wang, and J.~Wang.
\newblock Darc: Deep adaptive regularized clustering for histopathological image classification.
\newblock \emph{Medical image analysis}, 80:\penalty0 102521, 2022{\natexlab{a}}.

\bibitem[Li et~al.(2024)Li, Zhou, He, Cheng, and Huang]{fsam}
T.~Li, P.~Zhou, Z.~He, X.~Cheng, and X.~Huang.
\newblock Friendly sharpness-aware minimization.
\newblock In \emph{Proceedings of the IEEE/CVF conference on computer vision and pattern recognition}, pages 5631--5640, 2024.

\bibitem[Li et~al.(2022{\natexlab{b}})Li, Sun, and Tang]{DPC-FSC}
Y.~Li, L.~Sun, and Y.~Tang.
\newblock Dpc-fsc: An approach of fuzzy semantic cells to density peaks clustering.
\newblock \emph{Information Sciences}, 616:\penalty0 88--107, 2022{\natexlab{b}}.

\bibitem[Li et~al.(2023)Li, Hu, Wen, Evangelidis, Salahi, Wang, Tulyakov, and Ren]{EffienctVFormer}
Y.~Li, J.~Hu, Y.~Wen, G.~Evangelidis, K.~Salahi, Y.~Wang, S.~Tulyakov, and J.~Ren.
\newblock Rethinking vision transformers for mobilenet size and speed.
\newblock In \emph{Proceedings of the IEEE/CVF International Conference on Computer Vision}, pages 16889--16900, 2023.

\bibitem[Liang et~al.(2022)Liang, Ge, Tong, Song, Wang, and Xie]{EViT}
Y.~Liang, C.~Ge, Z.~Tong, Y.~Song, J.~Wang, and P.~Xie.
\newblock Not all patches are what you need: Expediting vision transformers via token reorganizations.
\newblock \emph{arXiv preprint arXiv:2202.07800}, 2022.

\bibitem[Liu et~al.(2021{\natexlab{a}})Liu, Dai, So, and Le]{gMLP}
H.~Liu, Z.~Dai, D.~So, and Q.~V. Le.
\newblock Pay attention to mlps.
\newblock \emph{Advances in neural information processing systems}, 34:\penalty0 9204--9215, 2021{\natexlab{a}}.

\bibitem[Liu et~al.(2021{\natexlab{b}})Liu, Lin, Cao, Hu, Wei, Zhang, Lin, and Guo]{swin}
Z.~Liu, Y.~Lin, Y.~Cao, H.~Hu, Y.~Wei, Z.~Zhang, S.~Lin, and B.~Guo.
\newblock Swin transformer: Hierarchical vision transformer using shifted windows.
\newblock In \emph{Proceedings of the IEEE/CVF international conference on computer vision}, pages 10012--10022, 2021{\natexlab{b}}.

\bibitem[Long et~al.(2023)Long, Zhao, Pi, Wang, and Wang]{BAT}
S.~Long, Z.~Zhao, J.~Pi, S.~Wang, and J.~Wang.
\newblock Beyond attentive tokens: Incorporating token importance and diversity for efficient vision transformers.
\newblock In \emph{Proceedings of the IEEE/CVF Conference on Computer Vision and Pattern Recognition}, pages 10334--10343, 2023.

\bibitem[Loshchilov and Hutter(2016)]{47}
I.~Loshchilov and F.~Hutter.
\newblock Sgdr: Stochastic gradient descent with warm restarts.
\newblock \emph{arXiv preprint arXiv:1608.03983}, 2016.

\bibitem[Loshchilov and Hutter(2017)]{46}
I.~Loshchilov and F.~Hutter.
\newblock Decoupled weight decay regularization.
\newblock \emph{arXiv preprint arXiv:1711.05101}, 2017.

\bibitem[Lotfi et~al.(2020)Lotfi, Moradi, and Beigy]{DPC-BDFN}
A.~Lotfi, P.~Moradi, and H.~Beigy.
\newblock Density peaks clustering based on density backbone and fuzzy neighborhood.
\newblock \emph{Pattern Recognition}, 107:\penalty0 107449, 2020.

\bibitem[Lu et~al.(2025)Lu, Chen, Ding, Tang, and Luo]{LWGANet}
W.~Lu, S.-B. Chen, C.~H. Ding, J.~Tang, and B.~Luo.
\newblock Lwganet: A lightweight group attention backbone for remote sensing visual tasks.
\newblock \emph{arXiv preprint arXiv:2501.10040}, 2025.

\bibitem[Ma et~al.(2024)Ma, Dai, Bai, Wang, and Fu]{StarNet}
X.~Ma, X.~Dai, Y.~Bai, Y.~Wang, and Y.~Fu.
\newblock Rewrite the stars.
\newblock In \emph{Proceedings of the IEEE/CVF Conference on Computer Vision and Pattern Recognition}, pages 5694--5703, 2024.

\bibitem[Maji et~al.(2013)Maji, Kannala, Rahtu, Blaschko, and Vedaldi]{aircraft}
S.~Maji, J.~Kannala, E.~Rahtu, M.~Blaschko, and A.~Vedaldi.
\newblock Fine-grained visual classification of aircraft.
\newblock Technical report, 2013.

\bibitem[Mooney(2018)]{BCCD}
P.~T. Mooney.
\newblock Blood cell images, 2018.
\newblock URL \url{https://www.kaggle.com/datasets/paultimothymooney/blood-cells}.
\newblock Accessed: 2025-04-09.

\bibitem[Nilsback and Zisserman(2008)]{flowers102}
M.-E. Nilsback and A.~Zisserman.
\newblock Automated flower classification over a large number of classes.
\newblock In \emph{2008 Sixth Indian conference on computer vision, graphics \& image processing}, pages 722--729. IEEE, 2008.

\bibitem[Pan et~al.(2021)Pan, Panda, Jiang, Wang, Feris, and Oliva]{IA-RED}
B.~Pan, R.~Panda, Y.~Jiang, Z.~Wang, R.~Feris, and A.~Oliva.
\newblock Ia-red2: Interpretability-aware redundancy reduction for vision transformers.
\newblock \emph{Advances in neural information processing systems}, 34:\penalty0 24898--24911, 2021.

\bibitem[Parkhi et~al.(2012)Parkhi, Vedaldi, Zisserman, and Jawahar]{OxfordIIIPet}
O.~M. Parkhi, A.~Vedaldi, A.~Zisserman, and C.~Jawahar.
\newblock Cats and dogs.
\newblock In \emph{2012 IEEE conference on computer vision and pattern recognition}, pages 3498--3505. IEEE, 2012.

\bibitem[Rao et~al.(2021)Rao, Zhao, Liu, Lu, Zhou, and Hsieh]{DynamicVIT}
Y.~Rao, W.~Zhao, B.~Liu, J.~Lu, J.~Zhou, and C.-J. Hsieh.
\newblock Dynamicvit: Efficient vision transformers with dynamic token sparsification.
\newblock \emph{Advances in neural information processing systems}, 34:\penalty0 13937--13949, 2021.

\bibitem[Rodriguez and Laio(2014)]{dpc}
A.~Rodriguez and A.~Laio.
\newblock Clustering by fast search and find of density peaks.
\newblock \emph{science}, 344\penalty0 (6191):\penalty0 1492--1496, 2014.

\bibitem[Sun et~al.(2016)Sun, Yang, Sun, and Wang]{SD-198}
X.~Sun, J.~Yang, M.~Sun, and K.~Wang.
\newblock A benchmark for automatic visual classification of clinical skin disease images.
\newblock In \emph{Computer Vision--ECCV 2016: 14th European Conference, Amsterdam, The Netherlands, October 11-14, 2016, Proceedings, Part VI 14}, pages 206--222. Springer, 2016.

\bibitem[Szegedy et~al.(2015)Szegedy, Liu, Jia, Sermanet, Reed, Anguelov, Erhan, Vanhoucke, and Rabinovich]{41}
C.~Szegedy, W.~Liu, Y.~Jia, P.~Sermanet, S.~Reed, D.~Anguelov, D.~Erhan, V.~Vanhoucke, and A.~Rabinovich.
\newblock Going deeper with convolutions.
\newblock In \emph{Proceedings of the IEEE conference on computer vision and pattern recognition}, pages 1--9, 2015.

\bibitem[Szegedy et~al.(2016)Szegedy, Vanhoucke, Ioffe, Shlens, and Wojna]{42}
C.~Szegedy, V.~Vanhoucke, S.~Ioffe, J.~Shlens, and Z.~Wojna.
\newblock Rethinking the inception architecture for computer vision.
\newblock In \emph{Proceedings of the IEEE conference on computer vision and pattern recognition}, pages 2818--2826, 2016.

\bibitem[Vaswani et~al.(2017)Vaswani, Shazeer, Parmar, Uszkoreit, Jones, Gomez, Kaiser, and Polosukhin]{transformer}
A.~Vaswani, N.~Shazeer, N.~Parmar, J.~Uszkoreit, L.~Jones, A.~N. Gomez, {\L}.~Kaiser, and I.~Polosukhin.
\newblock Attention is all you need.
\newblock \emph{Advances in neural information processing systems}, 30, 2017.

\bibitem[Veeling et~al.(2018)Veeling, Linmans, Winkens, Cohen, and Welling]{PCAM}
B.~S. Veeling, J.~Linmans, J.~Winkens, T.~Cohen, and M.~Welling.
\newblock Rotation equivariant cnns for digital pathology.
\newblock In \emph{Medical image computing and computer assisted intervention--mICCAI 2018: 21st international conference, granada, Spain, September 16-20, 2018, proceedings, part II 11}, pages 210--218. Springer, 2018.

\bibitem[Wang et~al.(2019)Wang, Ge, Lipton, and Xing]{imagentsketch}
H.~Wang, S.~Ge, Z.~Lipton, and E.~P. Xing.
\newblock Learning robust global representations by penalizing local predictive power.
\newblock \emph{Advances in neural information processing systems}, 32, 2019.

\bibitem[Wang et~al.(2021{\natexlab{a}})Wang, Yuan, Chen, Feng, and Yan]{PnP-DETR}
T.~Wang, L.~Yuan, Y.~Chen, J.~Feng, and S.~Yan.
\newblock Pnp-detr: Towards efficient visual analysis with transformers.
\newblock In \emph{Proceedings of the IEEE/CVF international conference on computer vision}, pages 4661--4670, 2021{\natexlab{a}}.

\bibitem[Wang et~al.(2021{\natexlab{b}})Wang, Xie, Li, Fan, Song, Liang, Lu, Luo, and Shao]{PVT}
W.~Wang, E.~Xie, X.~Li, D.-P. Fan, K.~Song, D.~Liang, T.~Lu, P.~Luo, and L.~Shao.
\newblock Pyramid vision transformer: A versatile backbone for dense prediction without convolutions.
\newblock In \emph{Proceedings of the IEEE/CVF international conference on computer vision}, pages 568--578, 2021{\natexlab{b}}.

\bibitem[Wei et~al.(2022)Wei, Fan, Xie, Wu, Yuille, and Feichtenhofer]{MaskedFeat}
C.~Wei, H.~Fan, S.~Xie, C.-Y. Wu, A.~Yuille, and C.~Feichtenhofer.
\newblock Masked feature prediction for self-supervised visual pre-training.
\newblock In \emph{Proceedings of the IEEE/CVF conference on computer vision and pattern recognition}, pages 14668--14678, 2022.

\bibitem[Wu et~al.(2021)Wu, Xiao, Codella, Liu, Dai, Yuan, and Zhang]{CVT}
H.~Wu, B.~Xiao, N.~Codella, M.~Liu, X.~Dai, L.~Yuan, and L.~Zhang.
\newblock Cvt: Introducing convolutions to vision transformers.
\newblock In \emph{Proceedings of the IEEE/CVF international conference on computer vision}, pages 22--31, 2021.

\bibitem[Xie et~al.(2021)Xie, Wang, Yu, Anandkumar, Alvarez, and Luo]{segformer}
E.~Xie, W.~Wang, Z.~Yu, A.~Anandkumar, J.~M. Alvarez, and P.~Luo.
\newblock Segformer: Simple and efficient design for semantic segmentation with transformers.
\newblock \emph{Advances in neural information processing systems}, 34:\penalty0 12077--12090, 2021.

\bibitem[Xie et~al.(2017)Xie, Girshick, Doll{\'a}r, Tu, and He]{ResNeXt}
S.~Xie, R.~Girshick, P.~Doll{\'a}r, Z.~Tu, and K.~He.
\newblock Aggregated residual transformations for deep neural networks.
\newblock In \emph{Proceedings of the IEEE conference on computer vision and pattern recognition}, pages 1492--1500, 2017.

\bibitem[Xu et~al.(2022)Xu, Zhang, Zhang, Sheng, Li, Dong, Zhang, Xu, and Sun]{EVO-ViT}
Y.~Xu, Z.~Zhang, M.~Zhang, K.~Sheng, K.~Li, W.~Dong, L.~Zhang, C.~Xu, and X.~Sun.
\newblock Evo-vit: Slow-fast token evolution for dynamic vision transformer.
\newblock In \emph{Proceedings of the AAAI conference on artificial intelligence}, volume~36, pages 2964--2972, 2022.

\bibitem[Yang and Newsam(2010)]{UCMerced}
Y.~Yang and S.~Newsam.
\newblock Bag-of-visual-words and spatial extensions for land-use classification.
\newblock In \emph{Proceedings of the 18th SIGSPATIAL international conference on advances in geographic information systems}, pages 270--279, 2010.

\bibitem[Yin et~al.(2022)Yin, Vahdat, Alvarez, Mallya, Kautz, and Molchanov]{A-ViT}
H.~Yin, A.~Vahdat, J.~M. Alvarez, A.~Mallya, J.~Kautz, and P.~Molchanov.
\newblock A-vit: Adaptive tokens for efficient vision transformer.
\newblock In \emph{Proceedings of the IEEE/CVF conference on computer vision and pattern recognition}, pages 10809--10818, 2022.

\bibitem[Yun et~al.(2019)Yun, Han, Oh, Chun, Choe, and Yoo]{44}
S.~Yun, D.~Han, S.~J. Oh, S.~Chun, J.~Choe, and Y.~Yoo.
\newblock Cutmix: Regularization strategy to train strong classifiers with localizable features.
\newblock In \emph{Proceedings of the IEEE/CVF international conference on computer vision}, pages 6023--6032, 2019.

\bibitem[Zeng et~al.(2022)Zeng, Jin, Liu, Qian, Luo, Ouyang, and Wang]{tcformer}
W.~Zeng, S.~Jin, W.~Liu, C.~Qian, P.~Luo, W.~Ouyang, and X.~Wang.
\newblock Not all tokens are equal: Human-centric visual analysis via token clustering transformer.
\newblock In \emph{Proceedings of the IEEE/CVF conference on computer vision and pattern recognition}, pages 11101--11111, 2022.

\bibitem[Zeng et~al.(2024)Zeng, Jin, Xu, Liu, Qian, Ouyang, Luo, and Wang]{TCFormerv2}
W.~Zeng, S.~Jin, L.~Xu, W.~Liu, C.~Qian, W.~Ouyang, P.~Luo, and X.~Wang.
\newblock Tcformer: Visual recognition via token clustering transformer.
\newblock \emph{IEEE Transactions on Pattern Analysis and Machine Intelligence}, 2024.

\bibitem[Zhang et~al.(2017)Zhang, Cisse, Dauphin, and Lopez-Paz]{43}
H.~Zhang, M.~Cisse, Y.~N. Dauphin, and D.~Lopez-Paz.
\newblock mixup: Beyond empirical risk minimization.
\newblock \emph{arXiv preprint arXiv:1710.09412}, 2017.

\bibitem[Zhao et~al.(2023)Zhao, Wang, Pan, Fan, and Lee]{DPC-FWSN}
J.~Zhao, G.~Wang, J.-S. Pan, T.~Fan, and I.~Lee.
\newblock Density peaks clustering algorithm based on fuzzy and weighted shared neighbor for uneven density datasets.
\newblock \emph{Pattern Recognition}, 139:\penalty0 109406, 2023.

\bibitem[Zhao and Sun(2025)]{CI2PViT}
X.~Zhao and Y.~Sun.
\newblock Compress image to patches for vision transformer.
\newblock \emph{arXiv preprint arXiv:2502.10120}, 2025.

\bibitem[Zhong et~al.(2020)Zhong, Zheng, Kang, Li, and Yang]{45}
Z.~Zhong, L.~Zheng, G.~Kang, S.~Li, and Y.~Yang.
\newblock Random erasing data augmentation.
\newblock In \emph{Proceedings of the AAAI conference on artificial intelligence}, volume~34, pages 13001--13008, 2020.

\end{thebibliography}
\end{document}